\documentclass[11pt]{article}
\pdfoutput=1
\usepackage[preprint]{acl}
\usepackage{float}
\usepackage{booktabs}
\usepackage{multirow} 
\usepackage{graphicx}
\usepackage{stfloats}
\usepackage{times}
\usepackage{latexsym}
\usepackage{footmisc}
\usepackage[T1]{fontenc}

\usepackage[utf8]{inputenc}

\usepackage{microtype}

\usepackage{inconsolata}

\usepackage{graphicx}

\title{MME-RAG: Multi-Manager-Expert Retrieval-Augmented Generation for Fine-Grained Entity Recognition in Task-Oriented Dialogues}

\author{
    Liang Xue$^{1,2}\thanks{Equal Contribution}$, 
    Haoyu Liu$^{1*}$, 
    Yajun Tian$^{1}$,
    Xinyu Zhong$^{2}$\\
    \textbf{Yang Liu$^{1}\thanks{Corresponding Author}$} \\
  $^{1}$Harbin Institute of Technology, $^{2}$Byering Technology \\
  \texttt{xueliang.xl@byering.com, liuhaoyu@stu.hit.edu.cn} \\
  \texttt{zzzzzxy@byering.com, liuyang@hit.edu.cn} \\
  }

\begin{document}
\maketitle
\begin{abstract}

Fine-grained entity recognition is crucial for reasoning and decision-making in task-oriented dialogues, yet current large language models (LLMs) continue to face challenges in domain adaptation and retrieval controllability. We introduce \textbf{MME-RAG}, a Multi-Manager-Expert Retrieval-Augmented Generation framework that decomposes entity recognition into two coordinated stages: \textit{type-level judgment} by lightweight managers and \textit{span-level extraction} by specialized experts. Each expert is supported by a \textbf{KeyInfo retriever} that injects semantically aligned, few-shot exemplars during inference, enabling precise and domain-adaptive extraction without additional training. Experiments on CrossNER, MIT-Movie, MIT-Restaurant, and our newly constructed multi-domain customer-service dataset demonstrate that MME-RAG performs better than recent baselines in most domains. Ablation studies further show that both the hierarchical decomposition and KeyInfo-guided retrieval are key drivers of robustness and cross-domain generalization, establishing MME-RAG as a scalable and interpretable solution for adaptive dialogue understanding.

\end{abstract}

\section{Introduction}

Task-oriented dialogue systems are essential in domains such as e-commerce, healthcare, and customer support\cite{dialog-systems}, where extracting key information from user utterances directly impacts task success. For instance, understanding user preferences enables personalized product recommendations, while accurately capturing symptoms facilitates reliable medical diagnoses. In these contexts, Named Entity Recognition (NER) plays a pivotal role\cite{NER}, as even minor extraction errors can cascade into suboptimal dialogue strategies and failed task completion. 

Despite the impressive capabilities of large language models (LLMs) \cite{deepseekai2025,openai2024gpt4,yang2025qwen3}, applying them to fine-grained entity recognition in complex dialogues remains challenging. Specifically: (i) domain-specific performance gaps, as general-purpose LLMs often underperform compared to fine-tuned models and may hallucinate entities \cite{yang-etal-2025-beyond} ; (ii) high adaptation costs, because domain transfer typically requires expensive fine-tuning and large-scale annotations; and (iii) limited retrieval relevance, since existing Retrieval-Augmented Generation (RAG) methods often return coarse-grained or noisy examples \cite{Lewis2020}. 

To address these challenges, we propose \textbf{MME-RAG} (Multi-Manager-Expert Retrieval-Augmented Generation), a framework that decomposes the NER task into lightweight type-level “judges” (managers) and expert boundary “solvers” (experts). Each expert leverages a \textbf{KeyInfo retriever}, which dynamically injects entity-centric, few-shot examples at inference time. This design enhances extraction accuracy, improves retrieval relevance, and ensures modularity and scalability without requiring costly model retraining.

In summary, our contributions are threefold:

\begin{enumerate}
\item We introduce a Multi-Manager-Expert architecture that decomposes entity extraction into judge-solve subtasks, improving controllability and scalability.  
\item We propose a KeyInfo-driven retrieval method for few-shot RAG, which enhances retrieval relevance and boosts extraction accuracy.  
\item We provide extensive empirical validation on public benchmarks and our newly introduced customer-service dataset, demonstrating that MME-RAG outperforms strong baselines in most cases.
\end{enumerate}

Extensive experiments confirm that MME-RAG achieves state-of-the-art performance in fine-grained entity recognition, highlighting the effectiveness of our retrieval-augmented, manager-expert decomposition strategy.

\begin{figure*}[t]
    \centering
    \includegraphics[width=1\linewidth]{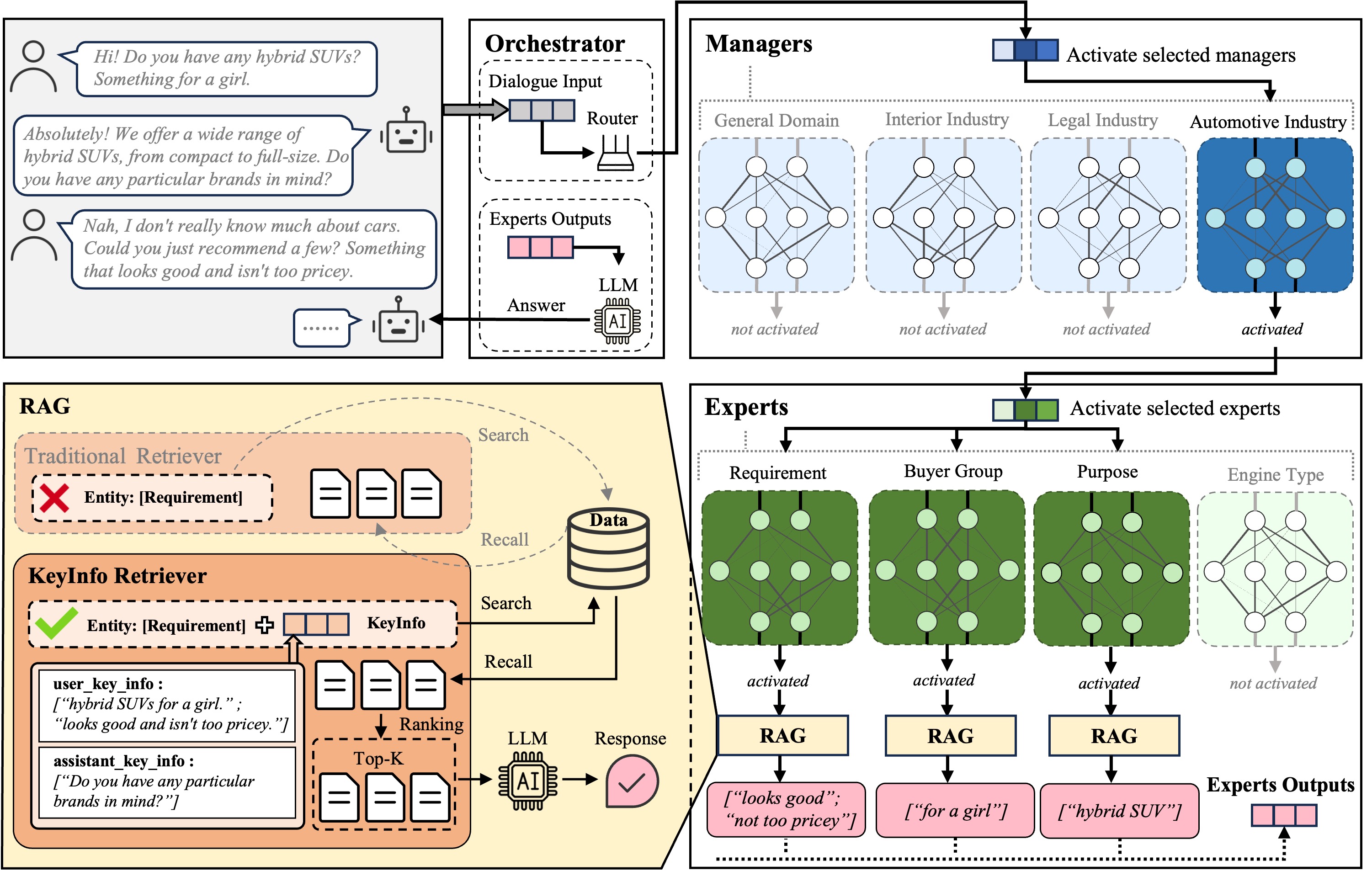}
    \caption{MME-RAG overview. The \textbf{Orchestrator} first routes user input to the appropriate domain \textbf{Managers} (e.g., automotive industry), which then activates relevant entity-level \textbf{Experts} (e.g., product type, requirement, buyer group). Each expert employs RAG to extract domain-specific entities, supported by the \textbf{KeyInfo Retriever} that enhances traditional retrieval by leveraging user- and assistant-level key information for fine-grained relevance ranking. The Database stores domain knowledge (entity definitions, history, and key info), enabling adaptive retrieval. }
    \label{fig:framework}
\end{figure*}

\section{Background and Related Work}

Entity recognition is a cornerstone of task-oriented dialogue systems, as accurately extracted entities guide dialogue policies and downstream decisions. 

\paragraph{Early Approaches.} Traditional NER began with rule-based systems \cite{rule-based} and feature-driven sequence labeling using HMMs and CRFs \cite{CRF}. Neural architectures, including BiLSTM-CRF \cite{BiLSTM-CRF} and CNN/RNN encoders \cite{CRF}, improved performance by automatically learning contextual features. Pre-trained encoders such as BERT \cite{BERT}, RoBERTa \cite{RoBERTa}, and DeBERTa \cite{DeBERTa} further advanced benchmark performance, yet remain data-hungry, brittle under domain transfer, and costly to extend to novel entity types.

\paragraph{LLM-Based NER.} Large language models (LLMs) recast NER as a generation task, enabling few-shot or zero-shot extraction through in-context learning \cite{In-context}. Approaches such as PromptNER \cite{PromptNER}, GPT-NER \cite{GPT-NER}, and UniversalNER \cite{UniversalNER} leverage prompting and type descriptions to improve generalization. Although these methods reduce annotation requirements, they still struggle with fine-grained subtype recognition, controllability, and computational cost in real-time dialogues.

\paragraph{Retrieval-Augmented NER.} RAG \cite{Lewis2020} grounds LLMs with external knowledge, mitigating hallucination. Models such as GPT-NER \cite{GPT-NER} and IF-WRANER \cite{IF-WRANER} demonstrate that retrieval can improve cross-domain extraction, while ColBERT \cite{ColBERT} provides token-level interactions for fine-grained relevance. Nevertheless, current RAG methods in dialogue are limited by shallow retrieval relevance, weak adaptability to emerging entity types, and tight coupling between type classification and boundary detection.

\paragraph{Summary.} In multi-turn, dynamic dialogues, these limitations are exacerbated by costly annotations, frequent schema evolution, and domain-specific semantics. Addressing them requires a retrieval framework that is fine-grained, adaptive, and seamlessly integrated with generative LLMs, motivating our MME-RAG approach.

\section{Methodology}

The proposed MME-RAG framework (Figure~\ref{fig:framework}) addresses fine-grained entity recognition in task-oriented dialogues by combining architectural decomposition with retrieval augmentation. It emphasizes precision, adaptability, and scalability, overcoming limitations of traditional NER pipelines and generic LLMs. The core concept is a Judge-Solve Paradigm, operationalized through a Multi-Manager-Expert architecture enhanced by a KeyInfo-driven retrieval mechanism.

\subsection{Judge-Solve Paradigm}

Our design draws inspiration from both computational complexity theory and cognitive science:

\paragraph{Proposition 1: Judgment is Easier than Generation.}  
Decision problems are generally less computationally intensive than constructive generation problems. For instance, Boolean satisfiability (SAT) is NP-complete when formulated as a decision problem, yet explicitly constructing a satisfying assignment is NP-hard \cite{Np}. Analogously, in natural language processing, determining whether a certain entity type appears in an utterance is simpler than generating its exact boundaries. Cognitive studies corroborate this notion: humans consistently perform recognition tasks more accurately and efficiently than generative reasoning tasks \cite{Axten1973HumanPS}. 

\paragraph{Proposition 2: Complex Problems Benefit from Decomposition.}  
Decomposing a complex task into smaller, manageable subproblems is a well-established paradigm in both algorithms and cognition. Divide-and-conquer and dynamic programming algorithms demonstrate substantial gains in efficiency and tractability \cite{Shyamasundar1996IntroductionTA,Kleinberg2005}. Cognitive problem-space theory similarly emphasizes hierarchical goal decomposition for tackling complex tasks \cite{Axten1973HumanPS}. 

Motivated by these principles, MME-RAG splits entity recognition into two stages: (1) lightweight \textbf{Managers} act as judges for type presence; (2) activated \textbf{Experts} perform precise boundary detection and value extraction.

\subsection{Multi-Manager-Expert Decomposition}

The MME architecture implements hierarchical and selective processing, as illustrated in Figure~\ref{fig:framework}. It is composed of three main layers:

\begin{itemize}
    \item \textbf{Orchestrator:} Receives dialogue input, routes through a dynamic router, coordinates Managers and Experts, and consolidates Expert outputs into the final response.
    \item \textbf{Managers:} Specialized in binary judgment for each domain (e.g., \textit{legal}, \textit{automotive}). Activation of a Manager triggers its Expert, reducing unnecessary computation and improving controllability.
    \item \textbf{Experts:} Once triggered, each Expert—a focused LLM-based generator—extracts entity boundaries for a narrow scope (e.g., \textit{engine type}). Restricting generation mitigates hallucination and enforces fine-grained distinctions.
\end{itemize}

This hierarchical decomposition, coupled with selective activation, enables scalability across domains with diverse and evolving entity schemas.

\subsection{KeyInfo-Driven Retrieval}

A central innovation of MME-RAG is the KeyInfo-driven retrieval mechanism, which addresses the limitations of coarse dialogue-level RAG. Traditional RAG methods often retrieve examples that are topically similar but semantically misaligned, introducing noise into few-shot prompts.

\paragraph{Data Layer.}  
The retrieval corpus is organized at the entity level, linking each target entity with its local dialogue context. Each example includes both user and assistant information, facilitating context-aware retrieval.

\paragraph{Retrieval Layer.}  
Rather than relying solely on embedding similarity, MME-RAG ranks examples based on two complementary signals: (i) \textit{user\_key\_info}, which captures salient entity mentions in the user utterances; and (ii) \textit{assistant\_key\_info}, which reflects system prompts that elicit or clarify user responses. By focusing retrieval on key expressions, the mechanism increases relevance, reduces irrelevant examples, and ensures semantic alignment between the retrieved instance and the target dialogue context.

\begin{figure}[H]
  \centering
  \includegraphics[width=1\linewidth]{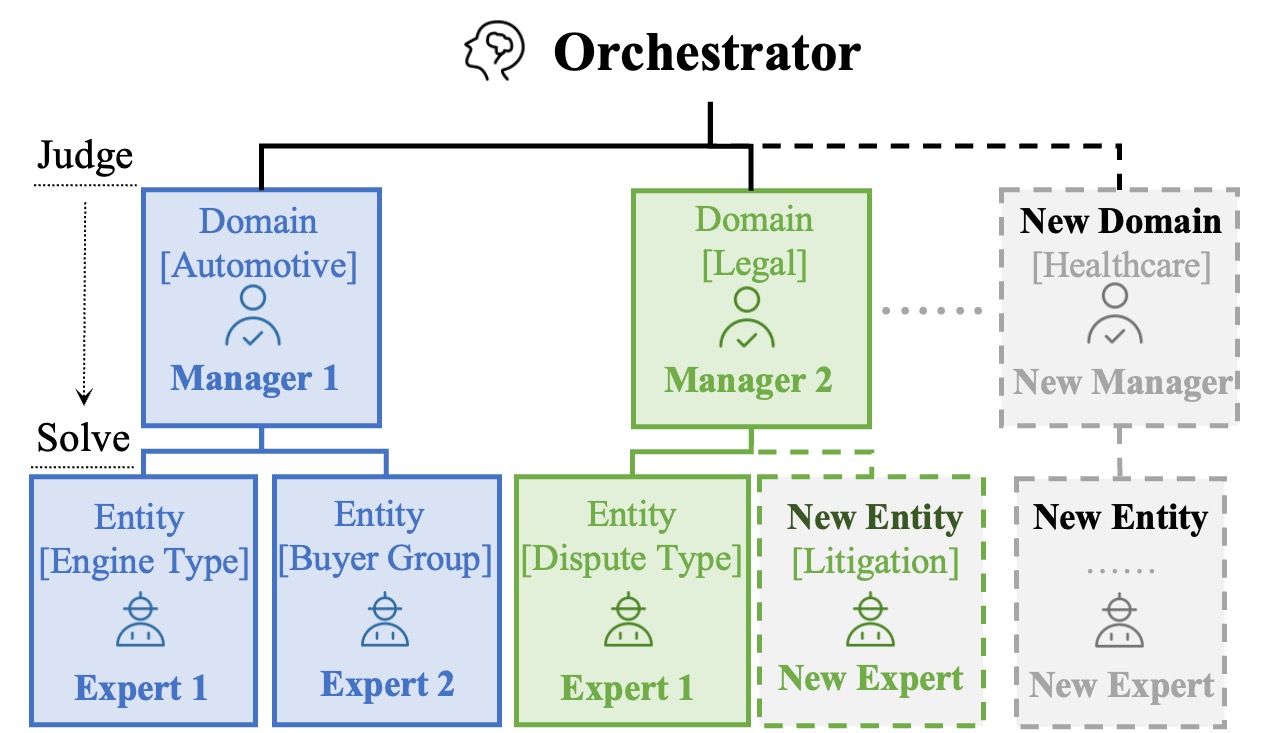}
  \caption{Illustration of MME-RAG’s modular expansion strategy. At the \textbf{Manager level}, new \textit{Managers} can be added for emerging domains (e.g., Healthcare). At the \textbf{Expert level}, new \textit{Experts} are introduced only when novel entity types arise (e.g., Litigation). This hierarchical design allows the system to scale across industries while minimizing disruption to existing modules, achieving cost-efficient and rapid domain–entity adaptation.}
  \label{fig:adaptation}
\end{figure}

\subsection{Rapid Domain and Entity Adaptation}

MME-RAG is designed for rapid adaptation to new domains and entity types without fine-tuning, leveraging both structural and knowledge-level modularity.

\paragraph{Structural Adaptation.}  
New Managers can be added for previously unseen domains, while new Experts are only required for novel entity types. For example, adding a \textit{Healthcare Manager} and a \textit{Litigation Expert} does not disrupt existing components such as the \textit{Automotive Manager} and its Experts (e.g., \textit{Engine Type, Buyer Group}). This ensures localized expansion, preserving system stability.

\paragraph{Knowledge-Level Adaptation.}  
Domain-specific examples, annotated with KeyInfo, are inserted into the retrieval corpus. During inference, these examples are dynamically injected into activated Experts’ prompts, enabling precise extraction without additional training. This approach allows MME-RAG to scale efficiently across industries and maintain robustness under evolving entity schemas.

\subsection{Customer-Service Dataset Construction}

\begin{figure}[t]
  \centering
  \includegraphics[width=1.0\linewidth]{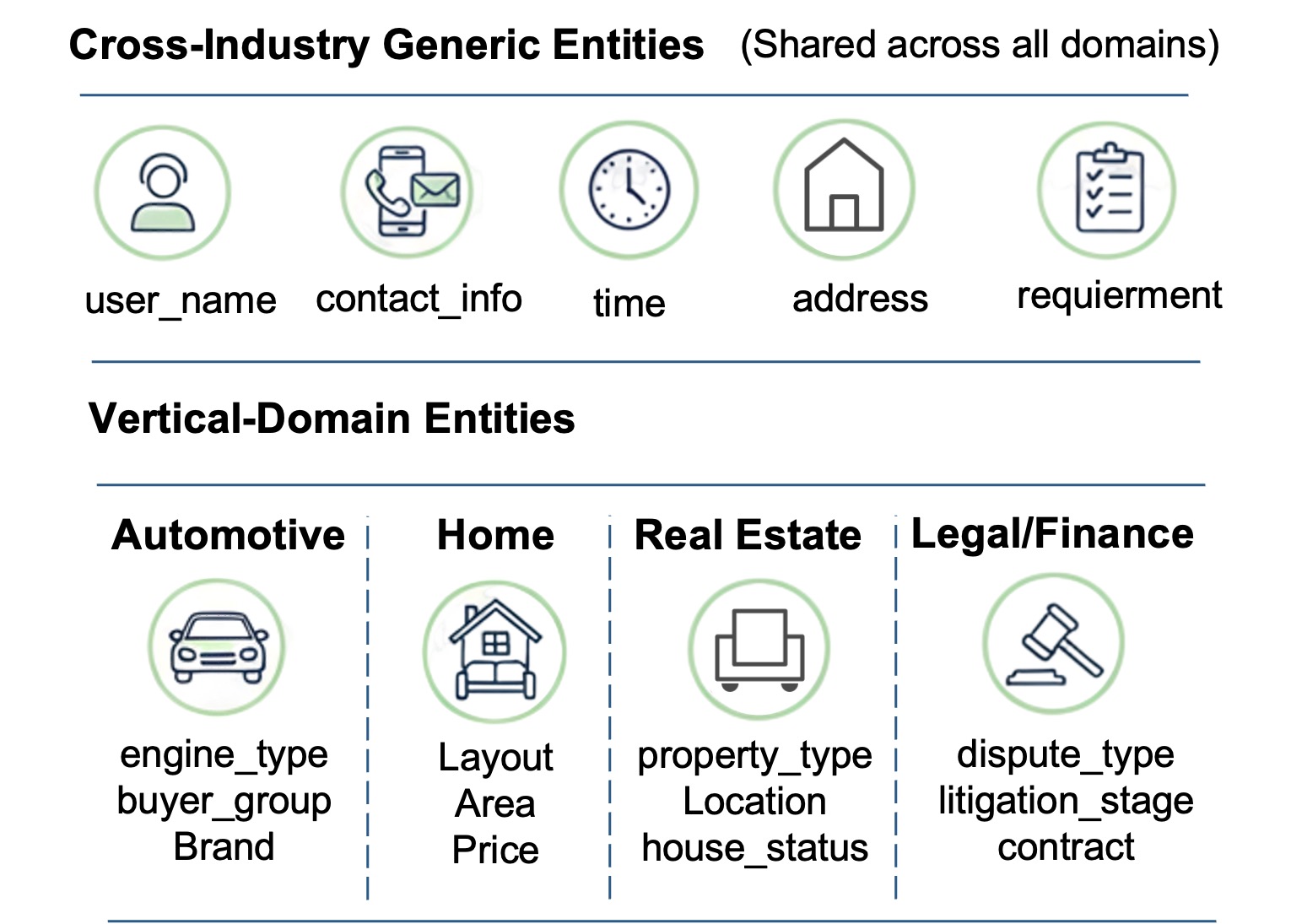}
  \caption{Overview of Customer-Service Dataset.The schema distinguishes between
    \textbf{Cross-Industry Generic Entities}---shared across all domains
    (e.g., user name, contact information, time, budget)---and
    \textbf{Vertical-Domain Entities}, specific to individual industries.
    Examples include automotive attributes (engine type, brand),
    home features (layout, price),
    real estate properties (type, location),
    and legal or financial aspects (dispute type, litigation stage).
    This fine-grained entity hierarchy facilitates precise domain understanding
    and robust entity-level reasoning.}
  \label{fig:datasets}
\end{figure}

\begin{figure}[h]
  \centering
  \includegraphics[width=1.0\linewidth]{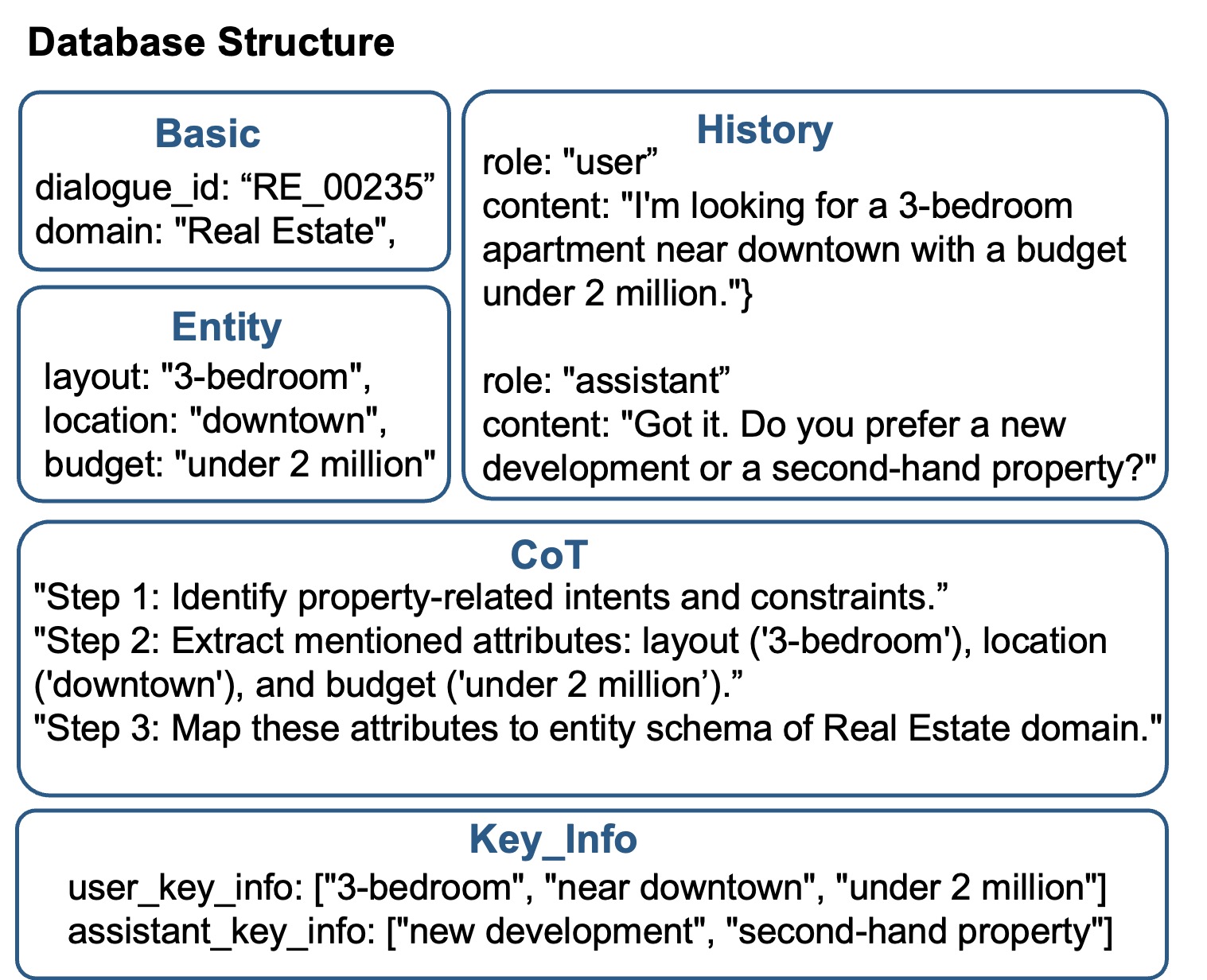}
  \caption{Illustration of a single annotated instance in the multi-domain dialogue dataset.Each record comprises five components:
    (1) \textbf{Basic Metadata} (dialogue ID, domain),
    (2) \textbf{Conversation History} (user--assistant turns),
    (3) \textbf{Entity Extraction} (e.g., layout, location, budget in the Real Estate domain),
    (4) \textbf{Chain-of-Thought (CoT)} reasoning for entity linking and intent derivation, and
    (5) \textbf{Key\_Info}, summarizing essential mentions from both participants.
    This unified annotation schema enables structured evaluation and supports
    KeyInfo-based augmentation for the MME-RAG framework.}
  \label{fig:database_structure}
\end{figure}

To assess MME-RAG under diverse domains and evolving schemas, we construct a multi-domain dialogue dataset designed around two principles: (i) entity-level alignment for retrieval-augmented reasoning and (ii) modular extensibility across domains and entity types. The dataset spans both cross-industry attributes and vertical-domain semantics, following a hierarchical taxonomy (Figure \ref{fig:datasets}). Each dialogue instance is annotated using a unified structure that integrates conversation context, reasoning signals, and retrieval-oriented cues, enabling precise supervision for both managers and experts (Figure \ref{fig:database_structure}). This design allows systematic study of domain adaptation and schema evolution. Additional statistics and annotation details are provided in Appendix~\ref{appendix:dataset}.

\section{Experiments}

\subsection{Experimental Setup}

\paragraph{Benchmarks.}  
We evaluate \textbf{MME-RAG} on three widely used public benchmarks and our newly constructed customer-service dataset.
\begin{itemize}
\item \textbf{CrossNER}\cite{CrossNER} spans five diverse domains (\textit{AI, Literature, Music, Politics, Science}) and contains fine-grained named entity categories, making it suitable for testing cross-domain generalization.
\item \textbf{MIT-Movie} and \textbf{MIT-Restaurant}\cite{MIT} are classic slot-filling benchmarks for dialogue-based entity recognition.
\item \textbf{Customer-Service Dataset}, our newly curated corpus, covers multiple real-world domains including \textit{general customer service}, \textit{automotive}, \textit{home}, \textit{real estate}, and \textit{legal} dialogues. Detailed dataset descriptions, statistics, and annotation schema are provided in Appendix~\ref{appendix:dataset}.
\end{itemize}

\begin{table*}[ht]
\centering
\setlength{\tabcolsep}{5pt}
\renewcommand{\arraystretch}{1.2}
\begin{tabular}{l|ccccc|cc}
\toprule
 & \multicolumn{5}{c|}{\textbf{CrossNER}} & \multicolumn{2}{c}{\textbf{MIT}}\\
 & AI & Literature & Music & Politics & Science & Movie & Restaurant\\
\midrule
InstructUIE & 48.40 & 48.80 & 54.40 & 49.90 & 49.40 & 63.00 & 20.99\\
UniNER      & 62.90 & 64.90 & 70.60 & 66.90 & 70.80 & 61.20 & 35.20 \\
GoLLIE      & 59.10 & 62.70 & 67.80 & 57.20 & 55.50 & 63.00 & 43.40 \\
KnowCoder   & 60.30 & 61.10 & 70.00 & 72.20 & 59.10 & 50.00 & 48.20 \\
GLiNER      & 57.20 & 64.40 & 69.60 & 72.60 & 62.60 & 57.20 & 42.90 \\
B2NER       & 59.00 & 63.70 & 68.60 & 67.80 & 72.00 & 67.60 & 53.30 \\
SaM         & 60.98 & 66.93 & 73.53 & 74.47 & 62.60 & 72.17 & 52.99 \\
IF-WRANER   & 68.81 & 75.52 & \textbf{85.43} & \textbf{79.80} & \textbf{75.31} & -- & -- \\
\midrule
\textbf{MME-RAG (Ours)} & \textbf{71.94} & \textbf{75.53} & 83.54 & 75.58 & 73.17 & \textbf{75.38} & \textbf{76.69} \\
\bottomrule
\end{tabular}
\caption{Performance comparison on CrossNER and MIT datasets. ``--'' indicates results not reported in the original paper and unavailable due to closed-source code.}
\label{crossner_mit}
\end{table*}

\paragraph{Metrics.}
We adopt task-specific metrics for different components.
For entity recognition, we report Precision (P), Recall (R), and F1 across datasets.
For retrieval, we use automatic metrics to measure:
(1) Relevance (Relevant / Irrelevant).
(2) Coverage (five buckets: 90–100\%, 60–90\%, 30–60\%, 1–30\%, 0).
(3) Semantic Similarity (a six-level scale: 1.0–0.0) for fine-grained alignment.

The six-level scale for semantic similarity is defined as follows:
\begin{itemize}
\item \textbf{1.0: Complete Match} — identical semantic meaning despite phrasing differences.
\item \textbf{0.8–0.9: Highly Similar} — nearly equivalent intent with minor variations.
\item \textbf{0.6–0.7: Moderately Similar} — partially overlapping semantics with shifted focus.
\item \textbf{0.4–0.5: Partially Similar} — some shared concepts but divergent intent.
\item \textbf{0.2–0.3: Low Similarity} — minimal conceptual overlap.
\item \textbf{0.0–0.1: Irrelevant/Opposite} — no or contradictory semantic relation.
\end{itemize}

\subsection{Entity Recognition on Benchmarks}

Table~\ref{crossner_mit} presents the results on CrossNER and MIT benchmarks. We compare MME-RAG—built upon the Qwen3-32B backbone—against strong and representative baselines, including InstructUIE, UniNER\cite{UniversalNER}, GoLLIE\cite{GoLLIE}, KnowCoder\cite{KnowCoder}, GLiNER\cite{GLiNER}, B2NER\cite{B2NER}, SaM\cite{SaM}, and the recent state-of-the-art \textbf{IF-WRANER}~\cite{IF-WRANER}.

MME-RAG consistently achieves superior performance across most datasets. Specifically, MME-RAG (71.94 F1) outperforms IF-WRANER (68.81 F1) in the AI domain of CrossNER and achieves comparable performance in Literature (75.53 vs. 75.52). While IF-WRANER shows stronger results in Music, Politics, and Science, MME-RAG significantly surpasses SaM on both MIT datasets (Movie: 75.38 vs. 72.17; Restaurant: 76.69 vs. 52.99).

Notably, our improvements are most significant in low-resource domains such as MIT-Restaurant, where MME-RAG achieves a substantial lead, underscoring the effectiveness of our KeyInfo-driven retrieval mechanism and multi-manager–expert decomposition for handling data sparsity and fine-grained entity diversity. Additionally, we compared MME-RAG with a BERT-based model implemented using PaddlePaddle~\cite{Paddle}, with detailed results presented in Appendix~\ref{appendix:Paddle}.

\subsection{Entity Recognition on Customer-Service Dataset}

To further validate the robustness and scalability of MME-RAG, we evaluate it on our newly constructed \textbf{Customer-Service Dataset}, which spans five practical domains: \textit{General}, \textit{Automotive}, \textit{Home}, \textit{Real Estate}, and \textit{Legal}.

We compare three configurations:
\begin{itemize}
\item \textbf{Baseline-8B}: A direct generation model based on a Qwen-8B backbone without decomposition or retrieval.
\item \textbf{+MME}: Incorporating the proposed multi-manager–expert decomposition on a Qwen-4B backbone.
\item \textbf{MME-RAG (Ours)}: Further integrating KeyInfo-driven retrieval on top of +MME, also utilizing a Qwen-4B backbone.
\end{itemize}

As shown in Table~\ref{custom_dataset}, MME-RAG (Qwen-4B) achieves the best overall performance across all five domains, consistently improving Precision, Recall, and F1 over the Baseline-8B and the +MME (Qwen-4B) variant. Notably, the combination of the multi-manager–expert decomposition and KeyInfo-driven retrieval yields substantial gains. For instance, in Real Estate, MME-RAG achieves an F1 of 99.48, significantly outperforming Baseline-8B's 93.83. Similarly, in the Legal domain, MME-RAG's F1 of 95.11 surpasses Baseline-8B's 93.44. These results confirm that both modules are indispensable for robust entity recognition and cross-domain generalization in customer-service dialogues, even when utilizing a smaller backbone model (Qwen-4B) compared to the baseline (Qwen-8B).

\begin{table*}[ht]
\centering
\setlength{\tabcolsep}{4pt}
\renewcommand{\arraystretch}{1.2}
\resizebox{\linewidth}{!}{
\begin{tabular}{l|ccc|ccc|ccc|ccc|ccc}
\toprule
 & \multicolumn{15}{c}{\textbf{Dataset (Metrics: P, R, F1)}} \\
 & \multicolumn{3}{c|}{General} & \multicolumn{3}{c|}{Automotive} & \multicolumn{3}{c|}{Home} & \multicolumn{3}{c|}{Real Estate} & \multicolumn{3}{c}{Legal} \\
\midrule
\multicolumn{16}{l}{\textbf{Qwen-8B}} \\
\midrule
Baseline          & 91.97 & 92.79 & 92.30 & 91.09 & 90.06 & 90.57 & 96.82 & 95.13 & 95.96 & 94.93 & 92.93 & 93.83 & 93.85 & 93.08 & 93.44 \\
\midrule
\multicolumn{16}{l}{\textbf{Qwen-4B}} \\
\midrule
+MME             & 94.79 & 93.16 & 93.93 & 91.58 & \textbf{93.24} & 92.38 & 96.39 & \textbf{98.19} & 97.27 & 95.63 & 98.01 & 96.74 & 94.41 & 94.53 & 94.44 \\
($\Delta$ P, $\Delta$ R, $\Delta$ F1)             & +2.82 & +0.37 & +1.63 & +0.49 & +3.18 & +1.81 & -0.43 & +3.06 & +1.31 & +0.70 & +5.08 & +2.91 & +0.56 & +1.45 & +1.00 \\
\midrule
+MME-RAG  & \textbf{95.75} & \textbf{95.46} & \textbf{95.55} & \textbf{93.85} & 92.62 & \textbf{93.23} & \textbf{98.14} & \textbf{98.19} & \textbf{98.17} & \textbf{99.55} & \textbf{99.41} & \textbf{99.48} & \textbf{95.40} & \textbf{94.85} & \textbf{95.11} \\
($\Delta$ P, $\Delta$ R, $\Delta$ F1)             & +3.78 & +2.67 & +3.25 & +2.76 & +2.56 & +2.66 & +1.32 & +3.06 & +2.21 & +4.62 & +6.48 & +5.65 & +1.55 & +1.77 & +1.67 \\
\bottomrule
\end{tabular}
}
\caption{Entity recognition results on the customer-service dataset across five domains. Each entry reports Precision (P), Recall (R), and F1 score.}
\label{custom_dataset}
\end{table*}

\subsection{Retrieval Performance}

We further evaluate the retrieval quality of MME-RAG from two complementary perspectives:
(1) comparison of different recall strategies, and
(2) semantic similarity analysis under various weighting schemes in the manager retrieval module.

\paragraph{Comparison of recall strategies.}
To assess how query formulation affects retrieval effectiveness, we compare three strategies:
(i) \textbf{Entity-level recall}, which uses the entire user dialogue context as the query;
(ii) \textbf{Dialogue-level recall}, which incorporates the latest user query, dialogue history, and agent replies; and
(iii) \textbf{KeyInfo-based recall}, which extracts concise user-side key expressions as the retrieval query.

\begin{table*}[h!]
\centering
\begin{tabular}{l|cc|ccccc}
\toprule
 & \multicolumn{2}{c|}{\textbf{Relevance}} & \multicolumn{5}{c}{\textbf{Coverage Distribution}} \\
 & Relevant & Irrelevant & 91--100\% & 61--90\% & 31--60\% & 1--30\% & 0\% \\
\midrule
Entity-level & 66.6\% & 33.3\% & 31.4\% & 4.6\% & 22.2\% & 8.3\% & 33.3\% \\
Dialogue-level & 62.0\% & 37.9\% & 32.4\% & 2.7\% & 13.8\% & 12.0\% & 37.9\% \\
KeyInfo-based & \textbf{83.3\%} & \textbf{16.6\%} & \textbf{47.2\%} & 4.6\% & 19.4\% & 12.0\% & \textbf{16.6\%} \\
\bottomrule
\end{tabular}
\caption{Evaluation of relevance and coverage under different recall strategies.}
\label{renovation_evaluation}
\end{table*}

As shown in Table~\ref{renovation_evaluation}, the KeyInfo-based method achieves the highest proportion of relevant samples (83.3\%) while minimizing irrelevant retrievals (16.6\%).
Moreover, it substantially increases the proportion of cases with full or near-full coverage (91–100\%), suggesting that focusing on essential user information yields more semantically aligned retrieval results compared to holistic or dialogue-level strategies.

\paragraph{Weighting strategies in the retrieval manager.}
We further examine the impact of different weighting configurations within the retrieval manager, focusing on the automotive domain. These weight ratios (e.g., 8–1–1) represent the relative importance of three dialogue components—user last reply, all user utterances, and all agent utterances—weighted as 8:1:1, respectively. A large-scale evaluation over 10,000 instances was conducted, measuring the semantic similarity between user queries and retrieved responses.

\begin{table*}[h!]
\centering
\begin{tabular}{lcccccc}
\toprule
\textbf{Weight Strategy} & \textbf{1.0} & \textbf{0.8--0.9} & \textbf{0.6--0.7} & \textbf{0.4--0.5} & \textbf{0.2--0.3} & \textbf{0.0--0.1} \\
\midrule
6–2–2 & 2.0\% & 20.0\% & 17.0\% & 12.0\% & 22.0\% & 27.0\% \\
6–3–1 & 2.0\% & 23.0\% & 16.0\% & 12.0\% & 19.0\% & 28.0\% \\
7–2–1 & 3.0\% & 26.0\% & 19.0\% & 12.0\% & 16.0\% & 24.0\% \\
7–1–2 & 4.0\% & 22.0\% & 22.0\% & 13.0\% & 20.0\% & \textbf{19.0\%} \\
8–1–1 & \textbf{4.0\%} & \textbf{28.0\%} & 18.0\% & 9.0\% & 19.0\% & 22.0\% \\
\bottomrule
\end{tabular}
\caption{Distribution of semantic similarity levels across different weighting strategies.}
\label{similarity_distribution}
\end{table*}

Table~\ref{similarity_distribution} presents the distribution of similarity levels across five weighting strategies.
The results indicate that the \textbf{8:1:1} configuration yields the highest proportion of strong semantic matches ($\ge 0.6$), reaching \textbf{50\%}, while the \textbf{7:1:2} setting achieves the best comprehensive recall rate ($\ge 0.2$) with \textbf{81\%}. These findings suggest that different weighting schemes can be tailored to optimize either high-precision retrieval or broader recall coverage.
These findings suggest that different weighting schemes can be tailored to optimize either high-precision retrieval or broader recall coverage.

Overall, these analyses confirm that MME-RAG’s retrieval module benefits substantially from KeyInfo-based query formulation and adaptive weighting strategies, both of which enhance semantic alignment and retrieval reliability across domains.

\section{Conclusion}

This work presented MME-RAG, a retrieval-augmented framework that decomposes fine-grained entity recognition into cooperative subtasks across a manager–expert hierarchy. Through the integration of KeyInfo-driven retrieval, MME-RAG effectively bridges few-shot generalization and domain adaptability without task-specific fine-tuning. Experiments across public and in-house benchmarks validate its superiority in robustness, interpretability, and efficiency. The proposed decomposition paradigm demonstrates that structured reasoning and targeted retrieval can jointly enhance the adaptability of LLM-based NER systems. Future research will explore reinforcement-based retrieval optimization and interactive feedback mechanisms to further improve adaptability in dynamic dialogue scenarios.

\section*{Limitations}

We acknowledge the following limitations of our work: (1) although the KeyInfo-driven retrieval mechanism improves relevance and robustness, it inevitably introduces additional inference latency, especially when the retrieval corpus scales up; (2) the effectiveness of MME-RAG depends on the coverage and quality of the retrieved examples—its performance may degrade in domains with scarce or noisy entity samples; (3) while the multi-manager–expert decomposition enhances controllability and interpretability, it also increases coordination complexity between modules, which may affect stability under domain shifts.

\bibliographystyle{unsrtnat}
\bibliography{acl_natbib}

\appendix
\section{Customer-Service Dataset Details}
\label{appendix:dataset}

The customer-service dataset is constructed to support fine-grained, industry-oriented entity recognition and reasoning. It spans multiple vertical domains and features unified annotation principles while preserving domain-specific distinctions. Figure~\ref{fig:datasets} provides an overview of the entity taxonomy, and Figure~\ref{fig:database_structure} illustrates the structure of a single annotated instance.

\paragraph{Domains.}
The dataset covers five major customer-service verticals:
\begin{itemize}
\item \textbf{General Customer Service}: routine metadata entities such as contact information, time expressions, budget ranges, and user-provided identifiers.
\item \textbf{Automotive}: comprehensive vehicle-related entities, including structured specifications (brand, sub-brand, series, model), numerical attributes (mileage, year, displacement, price), categorical properties (energy type, transmission, drivetrain, vehicle type), and purchasing–selling intent (intent vehicles vs. non-intent vs. selling vehicles). The schema also includes high-level purchasing conditions (e.g., usage purpose, target audience, broad constraints such as “low price” or “high cost-performance”).
\item \textbf{Home (Renovation)}: entities reflecting renovation needs across hard-decoration spaces, soft-decoration preferences, custom and non-custom furniture requirements, and style intentions. The schema additionally captures rejected or already-completed items, allowing contrastive reasoning about user constraints.
\item \textbf{Real Estate}: property-related entities including house type, floor/building type, area constraints, layout and room structure, and address normalization (province–city–district–town–street). Both intent and non-intent expressions are annotated when applicable.
\item \textbf{Legal / Finance}: entities describing dispute types, lawsuit and execution status, monetary amounts (normalized with comparison operators), evidence conditions, guarantee information, and social-security / housing-fund contribution status.
\end{itemize}

\paragraph{Entity Schema.}
To ensure cross-domain consistency, all entity types follow a unified structure with:
\begin{itemize}
\item \textbf{Class} (e.g., intent vs. non-intent vs. selling in Automotive; intent vs. non-intent in Real Estate building types; evidence/no-evidence in Legal),
\item \textbf{Surface Description} (user-provided expression),
\item \textbf{Normalized Values} (e.g., standardized brand--series–model; administrative divisions; numerical ranges with operators),
\item \textbf{Optional Sub-Entities} specific to the domain (e.g., layout structure, renovation style, lawsuit program stage).
\end{itemize}

\paragraph{Annotation Scope.}
Each dialogue instance contains: the full conversation history, CoT reasoning used for entity linking and constraint interpretation, summarized Key\_Info reflecting salient user and assistant mentions, aligned fine-grained entity annotations following the domain schema.

\paragraph{Detailed Taxonomies.}
The full specifications—including sub-entity definitions, value normalization rules, allowed category sets, and extended examples—are documented in \textbf{Appendix A}. They include:
\begin{itemize}
\item multi-level address normalization (province/city/district/town/street/community),
\item vehicle schemas with 20+ sub-entity types (year, mileage, energy, drivetrain, configuration, etc.),
\item renovation entities (hard/soft spaces, furniture, styles, construction state),
\item real-estate entities (house type, building type, layout structure, area constraints),
\item legal/financial entities (dispute type taxonomy, lawsuit phase, evidence types, monetary normalization, guarantee types, social-security status).
\end{itemize}

This dataset provides a challenging benchmark for fine-grained, domain-adaptive entity recognition under evolving schemas, and serves as the evaluation foundation for MME-RAG.

\section{Comparison with BERT-based PaddlePaddle Model}
\label{appendix:Paddle}

To further contextualize the performance of MME-RAG, we compare it with a BERT-based entity recognition model implemented using PaddlePaddle. The Paddle model follows a standard fine-tuning setup with token-level classification. Table~\ref{tab:paddle} reports precision, recall, and F1 scores across 18 automotive-related entity types from our customer-service dataset.

Overall, MME-RAG achieves consistently higher F1 scores across nearly all entity types, with particularly notable gains in challenging categories such as Vehicle Specs, Budget, and Purchase Requirements. These improvements highlight the advantages of our multi-manager–expert decomposition and KeyInfo-driven retrieval, especially for fine-grained and context-dependent entity types where pure encoder-based models tend to struggle. The results also demonstrate that MME-RAG remains robust even when compared to a domain-specific, fully fine-tuned encoder model.

\begin{table*}[ht] 
\centering
\begin{tabular}{l c c c c c c} 
\toprule
\multirow{2}{*}{\textbf{Entity Type}} & \multicolumn{3}{c}{\textbf{PADDLE(bert)}} & \multicolumn{3}{c}{\textbf{MME-RAG(ours)}} \\
\cmidrule(lr){2-4} \cmidrule(lr){5-7} 
& \textbf{p} & \textbf{r} & \textbf{f1} & \textbf{p} & \textbf{r} & \textbf{f1} \\
\midrule
Phone Number & 93.37\% & 99.41\% & 96.30\% & 97.47\% & 99.63\% & 98.54\% \\
WeChat ID & 98.17\% & 89.92\% & 93.86\% & 97.37\% & 92.50\% & 94.87\% \\
Province & 90.70\% & 100.00\% & 95.12\% & 96.59\% & 96.92\% & 96.75\% \\
City (multi-level) & 85.96\% & 79.03\% & 82.35\% & 90.75\% & 89.71\% & 90.23\% \\
Budget & 88.46\% & 71.88\% & 79.31\% & 82.25\% & 91.98\% & 86.85\% \\
Brand & 96.72\% & 93.65\% & 95.16\% & 96.50\% & 95.57\% & 96.03\% \\
Car Series & 82.50\% & 81.48\% & 81.99\% & 86.81\% & 87.41\% & 87.11\% \\
Car Model & 79.17\% & 79.17\% & 79.17\% & 83.05\% & 85.47\% & 84.24\% \\
Model Year & 91.67\% & 78.57\% & 84.62\% & 94.48\% & 93.84\% & 94.16\% \\
License Plate Date & 82.35\% & 87.50\% & 84.85\% & 94.12\% & 86.49\% & 90.14\% \\
Vehicle Specs & 70.27\% & 78.79\% & 74.29\% & 89.19\% & 84.62\% & 86.84\% \\
Energy Type & 92.31\% & 89.55\% & 90.91\% & 95.65\% & 92.96\% & 94.29\% \\
Transmission Type & 91.35\% & 81.90\% & 86.36\% & 90.38\% & 95.92\% & 93.07\% \\
Seat Count & 88.89\% & 88.89\% & 88.89\% & 96.30\% & 98.11\% & 97.20\% \\
Vehicle Type & 89.74\% & 92.11\% & 90.91\% & 92.13\% & 91.11\% & 91.62\% \\
Color & 97.87\% & 93.88\% & 95.83\% & 96.49\% & 96.49\% & 96.49\% \\
Purchase Requirements & 76.47\% & 83.87\% & 80.00\% & 87.10\% & 90.00\% & 88.52\% \\
Purchase Purpose & 74.55\% & 83.67\% & 78.85\% & 76.19\% & 94.12\% & 84.21\% \\
\bottomrule
\end{tabular}
\caption{Performance Comparison of Paddle (bert) and MME-RAG}
\label{tab:paddle}
\end{table*}

\end{document}